\documentclass{article}
\usepackage{graphicx}
\usepackage[utf8]{inputenc}
\usepackage{amsmath}
\usepackage{amsfonts}
\usepackage{cite}
\usepackage{url}
\usepackage{diagbox}
\usepackage{comment}
\usepackage{xcolor}
\usepackage{authblk}
\usepackage{float}

\title{Real time anomalies detection on videos}
\author{Fabien Poirier}
\affil{Paris 8 University, LIASD, France}

\begin{document}

\maketitle

\section{Abstract}

Nowadays, many places use security cameras. Unfortunately, when an incident occurs, these technologies are used to show past events. So it can be considered as a deterrence tool than a detection tool. In this article, we will propose a deep learning approach trying to solve this problematic. This approach uses convolutional models (CNN) to extract relevant characteristics linked to the video images, theses characteristics will form times series to be analyzed by LSTM / GRU models.
\vspace{0.5\baselineskip}

\textbf{Keywords:} Anomaly Detection, Video, Real-Time, VGG-GRU

\section{Introduction}

Currently, telesurveillance makes possible to control several places simultaneously thanks to a system of cameras, placed in a public or private space. The obtained images from these cameras are transmitted to a set of screens to be viewed and analyzed, then archived or destroyed. The goal of this surveillance is to control the security and safety conditions of these places. Actually, these images are analyzed by persons. So this task is very long and expensive. In addition, the effectiveness of this system depends on the attention and reactivity of the supervisor. With the development of artificial intelligence in many disciplines like image processing, audio processing, action recognition or anomaly detection. Could it be used to help the supervisor on his task or replace him ? To contribute to this task, any model must be able to detect quickly possible problems with maximum precision, or even to be able to predict them. We will broach the question if is possible to produce a real-time model for anomaly detection in video stream.

In this article, we will try to answer this question by proposing a deep learning approach. Due to is complexity, a video can be analyzed with three different ways, using only the sound, picture, or both. Since the majority of videos from surveillance cameras do not contain sound, which optimizes their storage, we will focus our research on image analysis. In addition, we will mainly focus on anomalies that may have a direct impact on the safety or security of people present in these videos. First of all we will present the past related work, in the second time we will introduce our architecture, next we will present our experiments and results, and we will conclude by explaining the future work.

\section*{Related Work}
Long Short-Term Memory (LSTM) neural networks are widely used for detecting anomalies in time series. Proposed by Hochreiter and Schmidhuber (1997) \cite{hochreiter1997long}, and improved by Gers et al. (2000) \cite{gers2000learning}, thanks to the introduction of a forgetting gate, the LSTM is a recurrent neural network (RNNs) model. RNNs were originally developed by Elman (1990)\cite{elman1990finding}, then adapted by Jordan (1997)\cite{jordan1997serial}. Unlike simple feed forward networks, they feature feedback links between units that allow them to memorize dynamic time series. Then in 2014 Du Tran et al. \cite{tran2015learning} proposed C3D (3D Convolution), an efficient model for detecting spatio-temporal criteria in the video stream capable to do actions or sport recognition. 3D convolution is based on successive image sequences in order to analyze the differences between them. This technique has shown very good results in many domains like human pose estimation, segmentation of medical images, actions recognition or even in video surveillance. Those performances are confirmed by the work of Waqas Sultani, Chen Chen, Mubarak Shah \cite{sultani2018real} or even those of Sabrina Aberkane, Mohamed Elarbi \cite{aberkane2019deep}. Due its 3rd dimension, C3D is a model allowing to extract relevant characteristics from movement. It is therefore generally used as a feature extractor for anomaly or violence detection in video streams. More recently, Majd and Safabakhsh (2019) \cite{majd2019motion} used a network called ConvLSTM (Convolutional Long Short Term Memory), combining a Convolution (CNN) for the feature extraction a very efficient technique on images and the memory of the LSTM in order to process images time series. This new motion-sensitive network is perfect for action recognition in video data. Finally, in 2020, several articles dedicated to these domains \cite{berenguer2019paired, ghani2019robust, peixoto2019toward} showed that these two architectures, C3D and ConvLSTM, are relevant. C3D performance seems superior to convLSTM, but this architecture is more complex to set up. In addition, the article of Nicola Convertini et al.\cite{convertini2020controlled} indicated that convLSTM model have also very good results.

\section{Approach}
\subsection{Preprocessing}
\begin{figure}[H]
\begin{center}
\includegraphics[width=11cm]{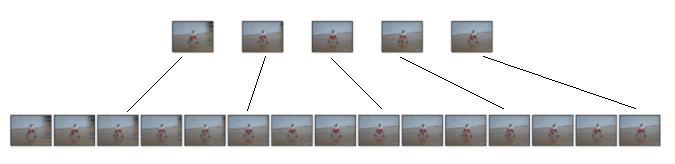}
\caption{Illustration of the video generator's functionality \cite{medium2024, github2024})} 
\label{Préparation}
\end{center}
\end{figure}

Most of the collected videos with an anomaly were cut to keep only this one. For each video, a sequence of x images was formed with a constant step between each image, so the extracted sequence covers the entire video (see figure \ref{Préparation}). Additionally, we augmented the data with zooms, crops, or mirror effects (horizontal flip). Data size was to large to load in memory, so we used a generator to realize this task.

\subsection{Architecture}
\begin{figure}[H]
\begin{center}
 \includegraphics[width=13cm]{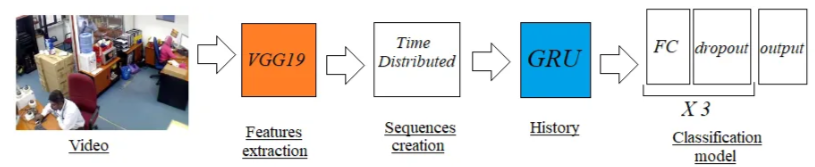}
 \caption{Proposed architecture} 
 \label{modele}
\end{center}
\end{figure}

\begin{figure}[H]
\begin{center}
\includegraphics[width=12cm]{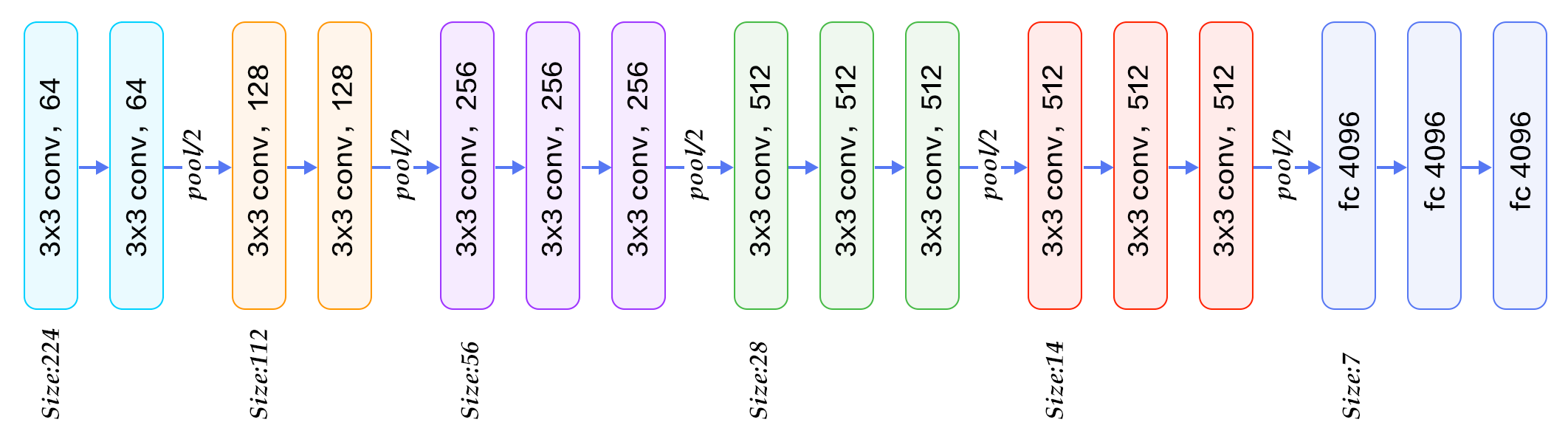}
\caption{VGG19 architecture \cite{surya2023efficacy}}  
\label{vgg19}
\end{center}
\end{figure}

The model proposed in figure \ref{modele} works on videos extract from dataset or streaming as a webcam. Before each analysis, it is possible to configure the desired prediction sequence (sequence expressed in seconds), the model will automatically calculate the step to apply between the images according to the number of fps of the video. Once this step has been calculated, the extraction of images can begin. For the feature extraction, we opted for a VGG model, more specifically VGG19. VGG is a model proposed by Simonyan and Zisserman (2014)\cite{simonyan2014very} recognized for these performances in the field of computer vision; he notably won the ILSVRC competition (ImageNet Large Scale Visual Recognition Challenge) in 2014 by achieving an accuracy of 92.7\% on the imageNet dataset. This model is composed of 23 layers grouped into 5 main blocks each composed of 2–4 convolution layers followed by a pooling layer (see figure \ref{vgg19}). In our architecture, we have partially re-trained it from the 12th layer representing the middle of the model and the 4th block. Since this convolution is 2D it is not suitable for image sequences. Each extracted image will be resized to a size of 112*112 then given one by one to the network in order to extract characteristics relevant. Regarding our batch, we have set it at 16. Once these features are extracted, this data will be passed to the Time Distributed layer. Unlike a 2D convolution with a loss of information due to a fusion of all our images to make them fit in a 2D format. The Time Distributed Layer will make it possible to accumulate the characteristics coming from different images by forming a prediction sequence. For each sequence, we fixed a size of 30 frames based on the current average (30 fps), the input data size of the Time Distributed layer is 30*112*112*3. The introduction of a GRU model to our architecture allows us to set up a history allowing us to better process our sequences. This type of model is composed of 2 doors: a forgetting door and an updating door. To begin, the data will pass through the forgetting gate, a gate that will allow to control the amount of information that must be forgotten and then the information kept, judged relevant are given to the update gate to be learned. Door which has the role of concatenate the new data with the previous state and then pass it through a sigmoid function to detect important components. Finally, in order to carry out our classification, we added to our model, a few full connected layers and a few dropout layers. Regarding the loss function given our number of classes we have chosen the binary cross-entropy function, but in the long term since certain anomalies can occur in parallel, we will opt for a categorical cross-entropy. For the optimizer we tested a wide variety of them, and the one with which we obtained the best results was SGD, but we had to reduce the learning coefficient.

\section{Experimention}
\subsection{Dataset}
Our dataset contains +4000 videos grouped into 13 classes among which we find: fights, fires, car accidents or videos without anomalies. So we have very varied scenes: day and night video, from different sources such as a telephone, surveillance camera, television camera, as well as different angles of view.

\subsection{Approach}
We have carried out numerous experiments along 3 main lines: defining a correct dataset, find an adequate preparation for our data and define the model to use. We started by training a model on all of our classes, but these data being massive, the learning time was impacted. Furthermore, the data being quite different from each other, the results obtained at present did not live up to our expectations. We therefore preferred to work on a sub-sample of our dataset compose of 2 classes: Fight/ Normal, which we have randomly separated in a 60/40 split. In the 2nd axis we should produce a single sequence summarizing the entire video, or several sequences formed from successive images. We have confronted these two approaches on videos where only the anomalies were kept and unmodified videos. The last axis of research was to find the optimal model. For this, we tested 5 differents convolutional networks such as vgg19, mobilenet and 3 personal convolutions. For the classification part, we also compared the performance of an LSTM and a GRU.

\section{Results}
\begin{figure}[H]
\begin{center}
\includegraphics[width=10cm]{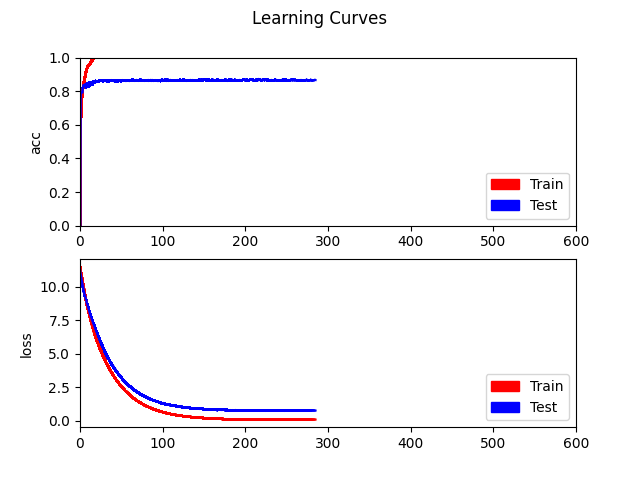}
\caption{Learning curves for the model Convolutional GRU}
\label{learning_curve}
\end{center}
\end{figure}

\begin{figure}[H]
\begin{center}
\includegraphics[width=10cm]{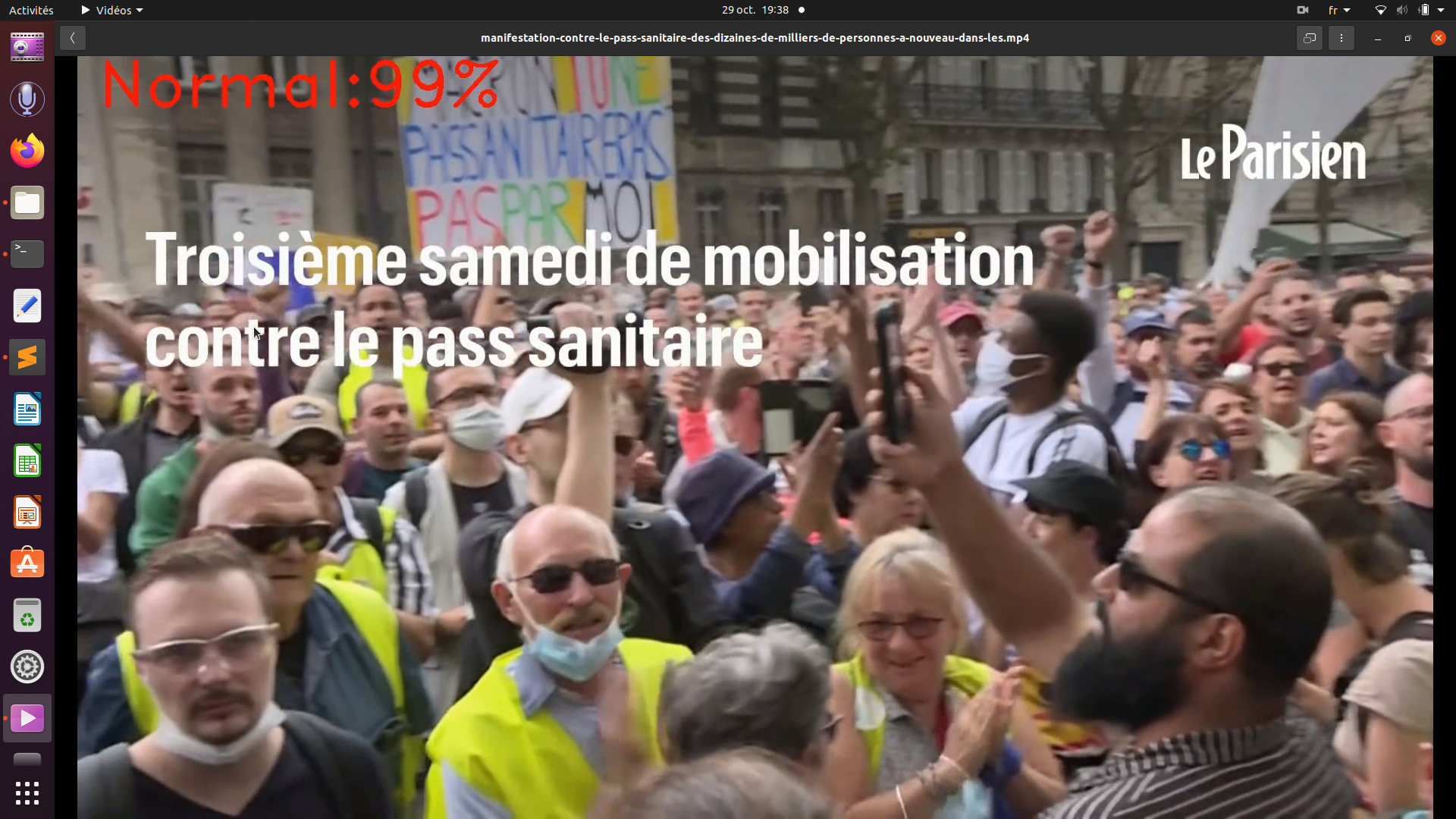}
\caption{example for the normal class detection.}
\label{normal_exemple}
\end{center}
\end{figure}

\begin{figure}[H]
\begin{center}
\includegraphics[width=10cm]{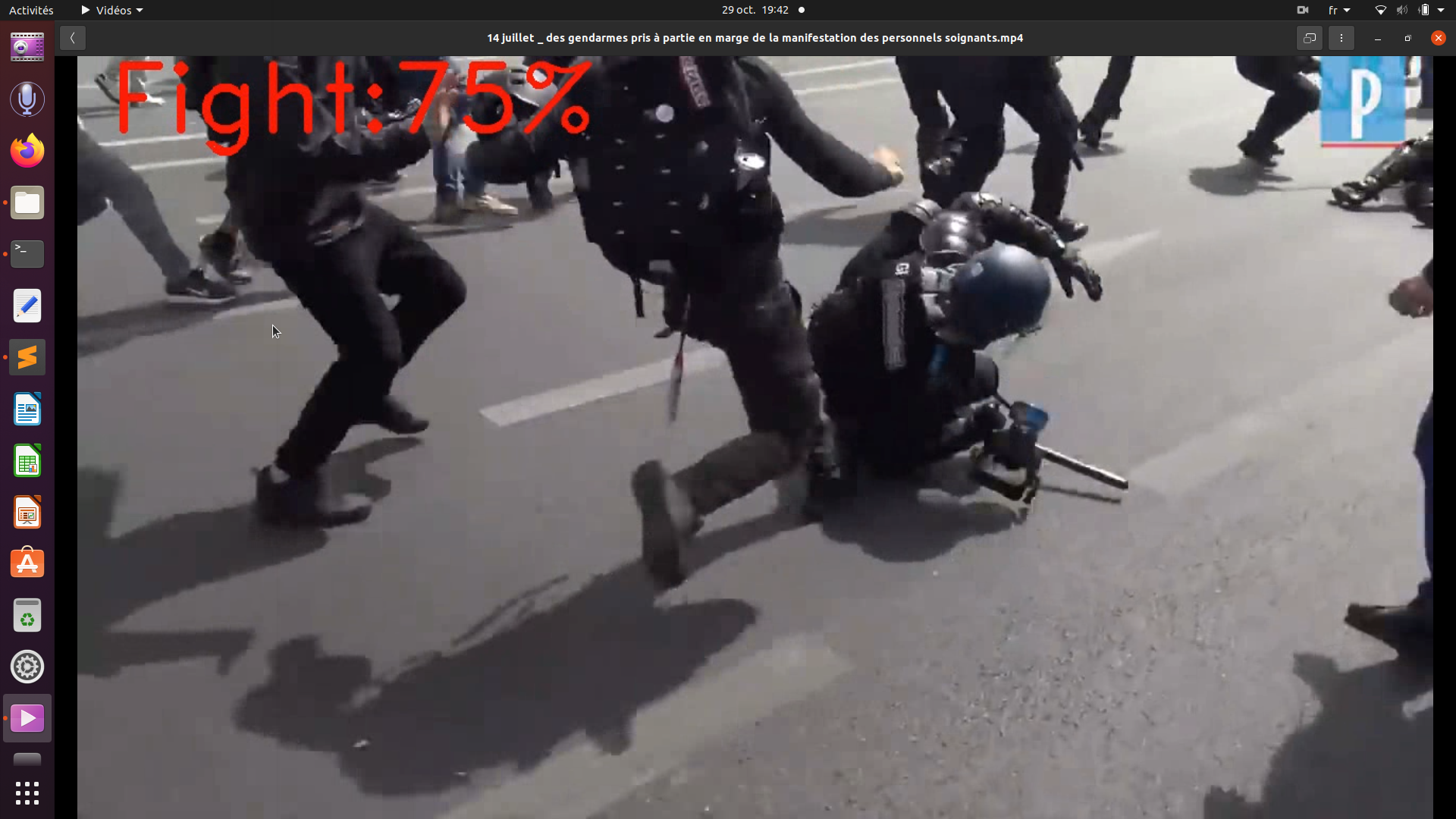}
\caption{Example for the fight class detection.}
\label{fight_exemple}
\end{center}
\end{figure}

\begin{figure}[H]
\begin{center}
\includegraphics[width=10cm]{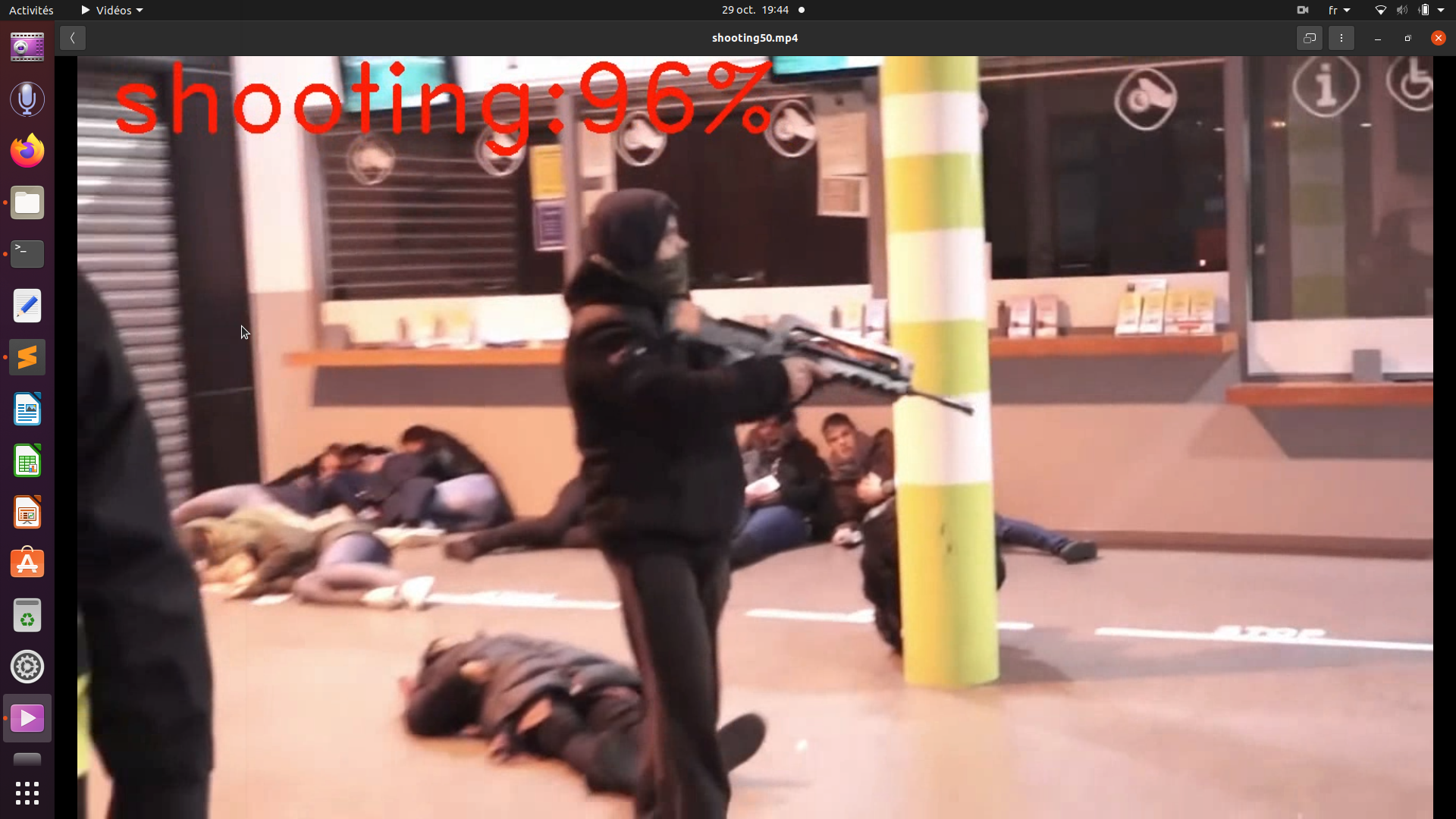}
\caption{Next step: shooting detection}
\label{shooting_exemple}
\end{center}
\end{figure}

In this section, we will start by presenting all the results obtained on our test set as a tables (see table \ref{tab_exemple}). Each tests was carried out on a set of finished videos, and only one prediction has been made.Additionally, to give an overview of the model's training performance, we provide the learning curve in figure \ref{learning_curve}, illustrating how the model's accuracy and loss evolved over the course of training. Then, we will show the performance of our approach on a videos against anti-sanitary pass (videos not included in the training and testing set), as illustrated in figures \ref{normal_exemple}, \ref{fight_exemple}, and \ref{shooting_exemple}.

\begin{table}[H]
 \begin{center}
 \caption{Result for differents model testing on the test set (A bloc represents 1 or 2 convolution layers + a pooling layer)} \label{tab_exemple}
   \tabcolsep = 1\tabcolsep
   \begin{tabular}{lcccc}
   \hline\hline\
    Model & Accuracy & Precision & Recall & F1-Score \\
   \hline
   Conv (3 blocs) + GRU + pred model & 78.8\% & 80\% & 75.9\% & 78.1\% \\
   Conv (5 blocs) + GRU + pred model & 85.1\% & 85.7\% & 84.2\% & 85\%  \\
   Conv (8 blocs) + GRU + pred model & 80.1\% & 78\% & 83.7\% & 80.7\% \\
   Mobilnet + GRU + pred model & 85.9\% & 86.9\% & 84.4\% & 85.7\% \\
   VGG19 + GRU + pred model & 86.5\% & 87.1\% & 85.7\% & 86.4\% \\
   VGG19 + LSTM + pred model & 85.1\% & 84.3\% & 86.3\% & 85.3\% \\
   VGG19 + GRU & 85\% & 86.1\% & 83.4\% & 84.7\% \\
   \hline
   \end{tabular}
 \end{center}
\end{table}

\section{Conclusion and future works}
In this article, we have shown that it is possible to use deep learning to detect in real time with a high precision anomaly coming from a video, such as fights. The proposed system was implemented using convolutional neuronal networks trained on images extracted from videos, combined with recurrent networks to analyze time series. The initial idea having been achieved, the main perspective is now to ensure that the model presented is able to process several types of anomalies.

Subsequently, it would be interesting to compare our approach to an approach using a C3D model. Finally, we can imagine a model coupled with object detection in order to improve these performances, or even develop a model capable of predicting anomalies a few seconds before they occur, which will facilitate the intervention of the persons concerned.

\bibliographystyle{plain}
\bibliography{references}

\begin{thebibliography}{10}

\bibitem{aberkane2019deep}
Sabrina Aberkane and Mohamed Elarbi.
\newblock Deep reinforcement learning for real-world anomaly detection in
  surveillance videos.
\newblock In {\em 2019 6th International Conference on Image and Signal
  Processing and their Applications (ISPA)}, pages 1--5. IEEE, 2019.

\bibitem{berenguer2019paired}
Abel~D{\'\i}az Berenguer, Meshia~C{\'e}dric Oveneke, Mitchel Alioscha-Perez,
  and Hichem Sahli.
\newblock Paired supervised learning and unsupervised pretraining of
  cnn-architecture for violence detection in videos.
\newblock In {\em BNAIC/BENELEARN}, 2019.

\bibitem{convertini2020controlled}
Nicola Convertini, Vincenzo Dentamaro, Donato Impedovo, Giuseppe Pirlo, and
  Lucia Sarcinella.
\newblock A controlled benchmark of video violence detection techniques.
\newblock {\em Information}, 11(6):321, 2020.

\bibitem{elman1990finding}
Jeffrey~L Elman.
\newblock Finding structure in time.
\newblock {\em Cognitive science}, 14(2):179--211, 1990.

\bibitem{github2024}
Patrice Ferlet.
\newblock keras-video-generators.
\newblock \url{https://github.com/metal3d/keras-video-generators}, 2019.

\bibitem{medium2024}
Patrice Ferlet.
\newblock Training a neural network with an image sequence — example with a
  video as input.
\newblock
  \url{https://medium.com/smileinnovation/training-neural-network-with-image-sequence-an-example-with-video-as-input-c3407f7a0b0f},
  2019.

\bibitem{gers2000learning}
Felix~A Gers, J{\"u}rgen Schmidhuber, and Fred Cummins.
\newblock Learning to forget: Continual prediction with lstm.
\newblock {\em Neural computation}, 12(10):2451--2471, 2000.

\bibitem{ghani2019robust}
Rana~Fareed Ghani et~al.
\newblock Robust real-time fire detector using cnn and lstm.
\newblock In {\em 2019 IEEE Student Conference on Research and Development
  (SCOReD)}, pages 204--207. IEEE, 2019.

\bibitem{hochreiter1997long}
Sepp Hochreiter and J{\"u}rgen Schmidhuber.
\newblock Long short-term memory.
\newblock {\em Neural computation}, 9(8):1735--1780, 1997.

\bibitem{jordan1997serial}
Michael~I Jordan.
\newblock Serial order: A parallel distributed processing approach.
\newblock In {\em Advances in psychology}, volume 121, pages 471--495.
  Elsevier, 1997.

\bibitem{majd2019motion}
Mahshid Majd and Reza Safabakhsh.
\newblock A motion-aware convlstm network for action recognition.
\newblock {\em Applied Intelligence}, 49(7):2515--2521, 2019.

\bibitem{peixoto2019toward}
Bruno Peixoto, Bahram Lavi, Jo{\~a}o Paulo~Pereira Martin, Sandra Avila, Zanoni
  Dias, and Anderson Rocha.
\newblock Toward subjective violence detection in videos.
\newblock In {\em ICASSP 2019-2019 IEEE International Conference on Acoustics,
  Speech and Signal Processing (ICASSP)}, pages 8276--8280. IEEE, 2019.

\bibitem{simonyan2014very}
Karen Simonyan and Andrew Zisserman.
\newblock Very deep convolutional networks for large-scale image recognition.
\newblock {\em arXiv preprint arXiv:1409.1556}, 2014.

\bibitem{sultani2018real}
Waqas Sultani, Chen Chen, and Mubarak Shah.
\newblock Real-world anomaly detection in surveillance videos.
\newblock In {\em Proceedings of the IEEE conference on computer vision and
  pattern recognition}, pages 6479--6488, 2018.

\bibitem{surya2023efficacy}
Janani Surya, Neha Pandy, Tyler~Hyungtaek Rim, Geunyoung Lee, MN~Swathi Priya,
  Brughanya Subramanian, Rajiv Raman, et~al.
\newblock Efficacy of deep learning-based artificial intelligence models in
  screening and referring patients with diabetic retinopathy and glaucoma.
\newblock {\em Indian Journal of Ophthalmology}, 71(8):3039--3045, 2023.

\bibitem{tran2015learning}
Du~Tran, Lubomir Bourdev, Rob Fergus, Lorenzo Torresani, and Manohar Paluri.
\newblock Learning spatiotemporal features with 3d convolutional networks.
\newblock In {\em Proceedings of the IEEE international conference on computer
  vision}, pages 4489--4497, 2015.

\end{thebibliography}

\end{document}